# Automated News Summarization Using Transformers

Anushka Gupta[1], Diksha Chugh[2], Anjum[3], Rahul Katarya[4]

[1,2,3,4] Delhi Technological University,New Delhi,India 110042
anushkag31@gmail.com,dikshachugh287@gmail.com,anjum_2792@yahoo.com,rahuldtu@gmail.com

**Abstract.** The amount of text data available online is increasing at a very fast pace hence text summarization has become essential. Most of the modern recommender and text classification systems require going through a huge amount of data. Manually generating precise and fluent summaries of lengthy articles is a very tiresome and time-consuming task. Hence generating automated summaries for the data and using it to train machine learning models will make these models space and time-efficient. Extractive summarization and abstractive summarization are two separate methods of generating summaries. The extractive technique identifies the relevant sentences from the original document and extracts only those from the text. Whereas in abstractive summarization techniques, the summary is generated after interpreting the original text, hence making it more complicated. In this paper, we will be presenting a comprehensive comparison of a few transformer architecture based pre-trained models for text summarization. For analysis and comparison, we have used the BBC news dataset that contains text data that can be used for summarization and human generated summaries for evaluating and comparing the summaries generated by machine learning models.

## 1 Introduction

The aim of news summarization is to create a concise summary from a long document or news articles such that no information is lost. In recent times, computing text summaries using Deep Learning has gained popularity

### 1.1 Need for Text Summarization

Automating summarization [1] would eliminate manual efforts. Shorter texts, which are summaries of longer texts, would reduce reading time. With the ever-growing amount of data, text summarization would reduce the size of files and hence solve the problem of storage. A shorter text or summary would provide more significant insights. Moreover, accurate summaries are very useful when it comes to text mining and data analysis.

## 1.2 Summarization Techniques

Text summarization can be broadly classified into two approaches [2] -
**Extractive Summarization** - In extractive summarization [3[], a summary from the given text is created by selecting a subset of the total sentence base. Most important phrases or sentences from the text are identified and selected based on a score that is computed depending on the words in that sentence.
**Abstractive Summarization** - In the method of abstractive summarization [4], an interpretation is first created by analysing the text document. Based on this interpretation, the machine predicts a summary. It transforms the text by paraphrasing sections of the original document.

This work will focus on abstractive summarization to create an accurate and fluent summary as this task is more challenging and simulates human perception for developing summaries. For this task, we have used some machine learning models pre-trained on a large dataset.

## 2 Related Work

The task of summarization using NLP first came into the picture in 1958. Initially, statistical approaches were used to compute a score for every sentence and then select the sentences with the highest scores. Several techniques were employed to calculate this score, such as TF-IDF [5], Bayesian models [6], etc. While these techniques were able to compute a sound summary by key phrase extraction, all of them were extractive approaches and were simply trimming the original text. Then the focus came onto utilizing Machine learning algorithms [1] for summarization, such as Bayesian Learning Models as was done in the paper [6]. These machine learning techniques proved to be successful for pattern recognition in texts and establishing a correlation between different words. In this section, we will be examining why and how machine learning techniques were employed for the task of summarization. Every text or sentence can be thought of as sequential data as the order of words is essential for the natural language interpretation and formation. In order to process sequential data, which is the case for most NLP problems, the architecture needs to retain information with the help of some memory.

One of the variants of RNN [7, 8] the LSTM network [9], retains sequential information with the help of connected nodes by keeping relevant information and forgetting insignificant information that helped in generating summaries. This methodology of LSTM network [9] was utilized to develop the encoder-decoder model [9]. Seq2seq models implemented with the help of the encoder-decoder framework for solving NLP tasks gave wonderful results, but there was still the issue of parallelization. Even though the sequential information is retained in the case of the encoder-decoder model [10] , the processing, in this case, is done by taking one input at a time as LSTM [8] takes only a single input at a time. This is a problematic situation as even though this model gives improved results, it proves to be unsuccessful for every possible case and defeats the purpose of creating machine perception.

This led to the addition of the Attention layer [8]. As depicted in Figure 1 [10], an attention layer in the encoder-decoder model [10] analyses the input sequence at every step, and based on the previous sequences, assigns a weight to it. The attention layer [11] creates vector matrices by considering every word in the sentence for one input. Hence, the attention layer forces the machine to look over the entire text as one input rather than separate sequences as separate inputs. This mechanism was extremely effective [11] for the abstractive approach and became popular. For this work, we have utilized the transformer architecture [11], developed by Google as a baseline model.

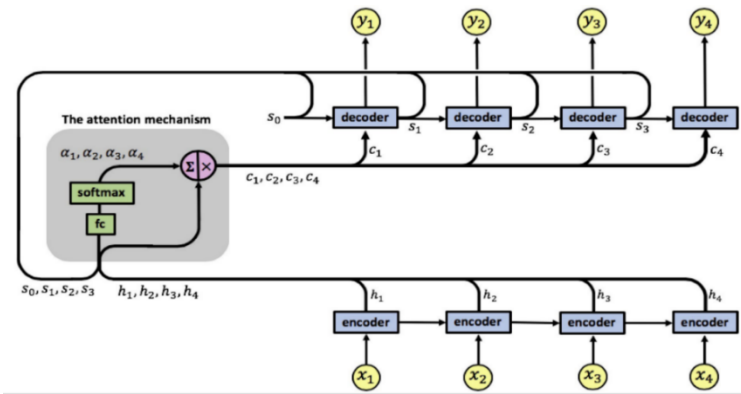

**Fig. 1.** Attention mechanism in Encoder-Decoder architecture [10]

The introduction of several pre-trained language models, for example, BERT [12], PEGASUS [13], UNiLM [14], GPT [15], etc., has transformed the field of NLP and their great results encouraged us to employ these technologies for the task of summarization. Most of the pre-trained models used in this project are based on Google's powerful Transformer architecture [11] that was developed in 2017. It is similar to RNN and is inspired from the encoder-decoder framework. Transformer model [11] was invented to solve natural language processing tasks that involved transforming an input sequence into an output sequence. These pre-trained language models from Hugging face library [16] can be used to solve multiple NLP problems.

## 3   Methodology

### 3.1 Dataset

From this section onwards, we will outline our basic experimental setup, discuss the evaluation metrics, and then describe various models that we used for our study. Then we will combine insights from our study and show the comparative performance of the models. The dataset we used was generated from a dataset used for text classification. It consists of 2225 BBC news website documents relating to stories used in the paper [17] in five topical areas from 2004-2005, all of whose rights, including copyright, are held by the BBC in the content of the original papers.

## 3.2 Preprocessing

This dataset consists of long news articles along with short summaries for comparison. The raw dataset was then cleaned using various pre-processing techniques such as:

**Lower casing** - To convert the input text into the same casing format so that all capital, lower case and mixed case are treated similarly.
**Eliminate Punctuation** - HTML tags and links- Removal of punctuations, links and tags that do not add meaning to the text such as "!"#$%&\'()*+,-./:;<=>?@[\\]^_{|}~`" to standardise the text.
**Eliminate Stopwords and frequently occurring words** - Removal of common words such as 'the', 'a', etc that are frequently used in a text but do not provide valuable information for downstream analysis.
**Stemming** - Reducing the inflected words to their root form.
**Lemmatization** - Reducing derived words to their base or root form while making sure that root words belong to the language.
**Contraction mapping** - Expanding the shortened version of words or syllables.

## 2.3 Model Explanation

**Basic Understanding of Transformer model -** The transformer network [11] is solely based upon multiple attention layers. It does not make use of RNN and is reliant on attention layers and positional encoding for remembering the sequence of words in the input sequence. The global dependencies created with the help of multiple attention layers help in creating parallelization in processing the input.

The transformer model [11] contains encoder and decoder layers, where each is connected to a multi-head attention layer and feed forward network layers. The model remembers the position and sequence of words with the help of cosine and sine functions that creates positional encoding. The multi-head attention layer [11] in the encoder and decoder layer applies a mechanism called self-attention. The input is fed into three connected layers to create query (Q), key (K), and value (V) vectors [11]. These vectors are split into n vectors.

$$Attention = softmax\left(\frac{QK^T}{\sqrt{d_k}}\right)V \quad (1)$$

Self-attention is applied on n separate vectors to create multi-head attention [11].

$$MultiHead(Q, K, V) = Concat(head_1, \dots, head_h)W^O \quad (2)$$

where, $head_i = Attention(QW_i^Q, KW_i^K, VW_i^V)$

Where the projections are parameter matrices [11]

$W_i^Q \in \mathbb{R}^{d_{model} \times d_k}$ , $W_i^K \in \mathbb{R}^{d_{model} \times d_k}$, $W_i^V \in \mathbb{R}^{d_{model} \times d_v}$ and $W^O \in \mathbb{R}^{hd_v \times d_{model}}$

Figure 2 [11] depicts the architecture of a transformer model. It contains an encoder and decoder layer and the various normalization and multi-head attention layers are also depicted in the figure.

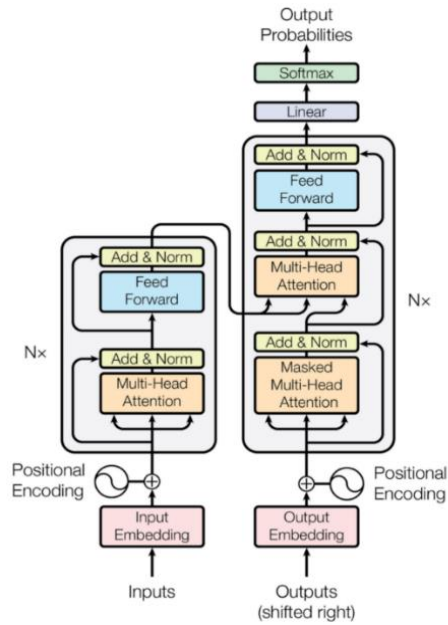

**Fig. 2.** Transformer Model Architecture [11]

**Pretrained Models based on Transformers -** Hugging face [16] works as an open-source for providing many useful NLP libraries and datasets. Its most famous library is the Transformer library. The transformer library consists of various pre-trained models to predict summaries of texts that can be fine-tuned for any dataset. Here we will be discussing some pre-trained models that were tuned and implemented for the BBC news dataset to give fairly good summaries. The models we used are as follows:

**Pipeline** – The pipelines are a great and quick way to use different pre-trained models for inference. These pipelines are objects that abstract most of the library's complicated code, offering a simple API dedicated to several tasks, including text summarization. Pipelines enclose the overall steps of every NLP process such as Tokenization, Inference, which maps every token into a more meaningful representation, and Decoding. The Hugging Face transformers summarization pipeline has made the task easier, faster and more efficient to execute in English language. We used the machine learning model that has been trained on the CNN news corpus by using a fine-tuned

BART algorithm [18] and is loaded from pipeline() using the task identifier: "summarization".

**BART** – BART stands for Bidirectional and Auto Regressive Transformers [18]. It is built with a seq2seq model trained with denoising as a pre-training purpose. It uses a standard seq2seq model architecture combining an encoder similar to BERT [12] and a GPT-like decoder [15]. The pre-training task involves changing the order of the original phrases randomly and a new scheme where text ranges are switched with a single mask token. The large model of BART [18] consists of twice as many layers as are present in the base model. It is quite similar to the BERT [12] model but BART contains about 10% more features than the BERT model of comparable size. BART's decoder is autoregressive, and it is regulated for generating sequential NLP tasks such as text summarization. The data is taken from the input but changed, which is closely related to the denoising pre-training objective. Hence, the input sequence embedding is the input of the encoder, and the decoder autoregressively produces output. We have used the "*facebook/bart-large-cnn*" pre-trained model and then the Bart tokenizer, which is constructed from the GPT-2 tokenizer. Hence words are encoded differently depending on their position in the sentence.

**T5** - T5 is the abbreviation for "Text-to-Text Transfer Transformer" [19]. The idea behind the T5 model is transfer learning [20]. The model was initially trained on a task containing large text in Transfer Learning before it was finely tuned on a downstream task so that the model learns general-purpose skills and information to be applied to tasks such as summarization T5 [19] uses a sequence-to-sequence generation method that feeds the encoded input via cross-attention layers to the decoder and generates the decoder output autoregressive. We have fine-tuned a T5 model [19], where the encoder takes an input a series of tokens which are mapped to a sequence of embeddings. A block containing two subcomponents are present in the encoder block namely, a self-attention layer and feed forward network. The decoder and encoder are similar in structure, except that there's a generalized attention mechanism after every self-attention layer. This allows the model to operate only on the previous outputs. The final decoder block produces an output which is fed into another layer. This final layer is a dense layer where the activation function is softmax. The weights from the output of this layer are fed into the input embedding matrix.

**PEGASUS** - PEGASUS stands for Pre-training with Extracted Gap-sentences for Abstractive Summarization Sequence-to-sequence models [13]. In this model, significant lines are eliminated from the input text and are compiled as separate outputs. Also, choosing only relevant sentences outperforms the randomly selected sentences. This methodology is preferred for abstractive summarization as it is similar to the task of interpreting the entire document and generating a summary. It is used to train a Transformer model on a text data resulting in the PEGASUS model. The model is pre-trained on CNN/DailyMail summarization datasets.

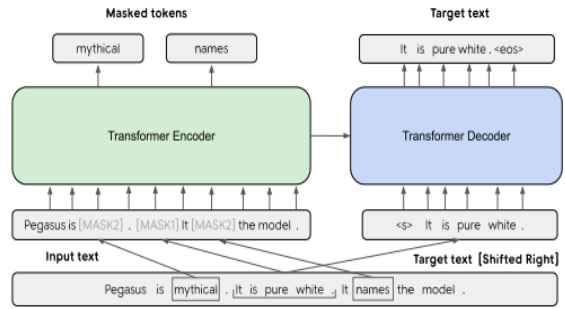

**Fig. 3.** The PEGASUS Model Architecture

As shown in Figure 3 [13], it was found that it works as a prior-training aim for text summarization to mask complete sentences from the text and create gap-sentences from the remaining document.

## 4 Results

**4.1 Qualitative Analysis**

For generating summaries, we fine-tuned the following transformer-based pre-trained language models from the Hugging face library [16]. The BBC News Dataset was used for generating summaries that consisted of text and human generated summaries which are summaries that have been written by humans.

We used the human generated summaries to perform a comprehensive analysis of the summaries generated by different models. Table 1 gives the results obtained from different models which have been compared with the given reference summary - "*An incident of robbery that occurred at the shopping complex last night was reported at the local police station this morning. A lot of valuables were stolen and multiple such robberies have been reported in that area. The people have been asked to stay alert and notice any suspicious activity. A CCTV camera from a nearby house captured the incident and there were a total four robbers who can be seen carrying bags. The shopkeeper suffered a loss and hopes the police catch the culprits*.".

**Table 1.** Comparison of summaries generated by models

| Models | Summaries generated by models |
|---|---|
| Pipeline - BART | "Last night, an incident of a robbery was reported at the shopping complex. According to the shopkeeper, as he reached his shop in the morning, he found that the door was already open and many valuable items were stolen from the shop. They broke into the shop and within 30-40 min came back with bags full of valuable items." |

| BART modified | "Last week an incident of a robbery was reported at the shopping complex. The shopkeeper who has a shop in the complex said this incident to the local police station in the morning. The police suspect that these robberies are somehow the shopkeepers association has issued a notice and asked everyone in the" |
|---|---|
| T5 | "shopkeeper reported robbery at shopping complex in the morning. he filed a complaint with police and hopes they catch the culprits. the shopkeepers association has issued a notice and asked everyone to stay alert.." |
| PEGASUS | "The shopkeeper who has a shop in the complex said this incident to the local police station in the morning as they found out about the robbery in the according to the as he reached his shop in the he found that the door was already open and many valuable items were stolen from the he immediately contacted the nearby police station and filed a the shopkeeper hopes that the police catch the culprits as soon as possible so that he can recover the items that were stolen from his there have been multiple such complaints in the past few weeks in this and the police suspect that these robberies are somehow the shopkeepers association has issued a notice and asked everyone in the area." |

**Pipeline-** The summaries generated by the pipeline model included sentences that deviated the most compared to the original reference summary and focused on unimportant sentences.

**BART-** BART gave better results than the pre-trained Bart model in the pipeline method. The summaries generated were fluent, accurate, and integrated supporting evidence from the input document. Hence, the summaries generated by the BART pre-trained model demonstrate that the BART model is useful for text understanding.

**T5** - The T5 model shows good results and a comparatively higher Rogue score F-value. The summaries generated are coherent and accurate. The text meaning was preserved in these summaries and aligned well with the original summary.

**PEGASUS** - On fine-tuning and using this model for our dataset, we got fluent and coherent summaries. These summaries were of high linguistic quality and closely matched the style of ground truth summaries. But the results show that the summaries generated were too short and incomplete.

### 4.2 Quantitative Analysis

We have compared a variety of different approaches for text summarization tasks while keeping as many factors constant as possible as shown in Table 2. From Table 2 we can conclude that T5 outperformed all other models by comparing the Rouge scores for each model. The Pipeline model on the other hand has the least ROUGE score value

[21], hence the summaries generated by using Transformer's pipeline model gives the least acceptable result.

Table 2. Evaluation and Comparison of mean ROUGE Scores

| Models | Evaluation Metrics | | |
| --- | --- | --- | --- |
| | *ROUGE-1* | *ROUGE-2* | *ROUGE-L* |
| Pipeline - BART | 0.38 | 0.28 | 0.38 |
| BART modified | 0.40 | 0.28 | 0.40 |
| **T5** | **0.47** | **0.33** | **0.42** |
| PEGASUS | 0.42 | 0.29 | 0.40 |

## 5. Conclusion

We implemented pre-trained language models, which were based upon the transformer architecture for the task of summarization. We concluded from our research that finely tuned transformers based pre-trained language models gave wonderful results and created a sound and fluent summary for a given text document. We computed ROUGE scores [21] for the predictions made by each of the models for comparative studies and concluded that the T5 model outperformed all other models.

Future work should focus on building more robust models. We can further extend our algorithm to create summaries of variable length and also apply this to multi-document summarization. A hybrid of the above models can be used to improve the accuracy, fluency, and coherence of the summaries.